\documentclass[times, review, 10pt]{elsarticle}
\usepackage[top=4.3cm, bottom=4.3cm, left=4.8cm, right=4.8cm]{geometry}

\usepackage{amssymb}
\usepackage{amsmath}
\usepackage{xcolor}
\usepackage{amssymb}
\usepackage{amsmath}
\usepackage{multirow}
\usepackage{bbding}

\usepackage{marvosym}
\usepackage{booktabs}
\usepackage[ruled,linesnumbered]{algorithm2e}

\usepackage{tabularx}
\usepackage{graphicx}
\usepackage{url}

\journal{Pattern Recognition}

\begin{document}

\begin{frontmatter}

\title{\fontsize{14pt}{16.8pt}\selectfont SpikeTAD: Spiking Neural Networks for End-to-End Temporal Action Detection}

\author{\fontsize{8pt}{9.6pt}\selectfont
  Min Yang\textsuperscript{a}, \sep
  Mi Zhou\textsuperscript{a}, \sep
  Limin Wang\textsuperscript{a,\Letter} 
}

\affiliation{organization={$^a$State Key Laboratory for Novel Software Technology, Nanjing University},
            city={Nanjing},
            state={Jiangsu},
            country={China}}

\begin{abstract}
Video understanding is a crucial part of computer vision, with numerous application scenarios. With the increasing popularity of mobile devices, an increasing number of efforts are trying to deploy video understanding models on them.
However, existing video understanding models are difficult to deploy due to their large size and prohibitive power consumption. Spiking Neural Networks (SNNs) have shown bioplausibility and low power advantages over Artificial Neural Networks (ANNs), especially on neuromorphic chips which are regarded as essential components of future mobile devices. However,
excessively long conversion time-steps and severe performance degradation problems limit
their application. 
To solve the problems above, we explore the application of SNNs on temporal action detection (TAD), which is an important task in video understanding, and propose the first SNN-based end-to-end TAD architecture coined as \textbf{SpikeTAD}. While maintaining extremely low power consumption, SpikeTAD achieves an average mAP of 67.2\%
in THUMOS14 and 37.42\% in ActivityNet-1.3, demonstrating the feasibility of a low-power TAD model. Our code is available at \url{https://github.com/MCG-NJU/SpikeTAD}.
\end{abstract}


\end{frontmatter}

\pagenumbering{arabic}

\section{Introduction}
The video understanding community continues to evolve, with the research focus gradually shifting from short-term video action classification to long-term video content comprehension. As an important basic task in the field of video understanding, Temporal Action Detection (TAD) refers to localizing and classifying action snippets in untrimmed long videos. Existing work~\cite{actionformer,tadtr,bmn,vittad} continues to advance the detection performance, allowing the TAD model to be applied to a wider range of scenarios~\cite{thumos14,anet}. However, these TAD works remain focused on improving the detection results, with mobile device deployment not being a priority in current TAD research. Although some methods~\cite{pbd} attempt to explore how to build efficient TAD models for limited computing resources, they remain focused on GPU-based inference, without considering the power consumption which is crucial for mobile devices. 

Spiking neural networks (SNNs)~\cite{snn} are biologically plausible neural networks based on the dynamic characteristics of biological neurons. As the third generation of artificial neural networks, SNNs have attracted great attention due to their low-power advantages
over artificial neural networks (ANNs). Spiking neurons inside SNNs emit spikes when the membrane potentials exceed a threshold, and trigger
sparse additions only when they receive a spike. From the perspective of computational hardware operations, multiplication operations in state-of-the-art deep
ANNs are replaced by
accumulation (AC) operations in SNNs, achieving lower energy consumption compared to ANNs. However, the negative impact of complex neuronal dynamics and spike-driven nature is that SNNs are difficult to train and have limited task performance and application scenarios. Most SNN methods~\cite{snnresnet,spikformer} try to reduce the time-step and improve SNN to obtain efficient and powerful models, but they are all limited to simple classification tasks.  

Applying SNNs to TAD models on neuromorphic chips faces much greater challenges than image models. First, TAD requires further understanding of video semantics, which is more complex than images. 
Second, videos have additional time dimension compared with images, and SNN requires an additional time-step for spike accumulation, making the power consumption of the model greater than that of image models. 

\begin{figure}[ht]
	\centering
	\includegraphics[width=1.0\columnwidth]{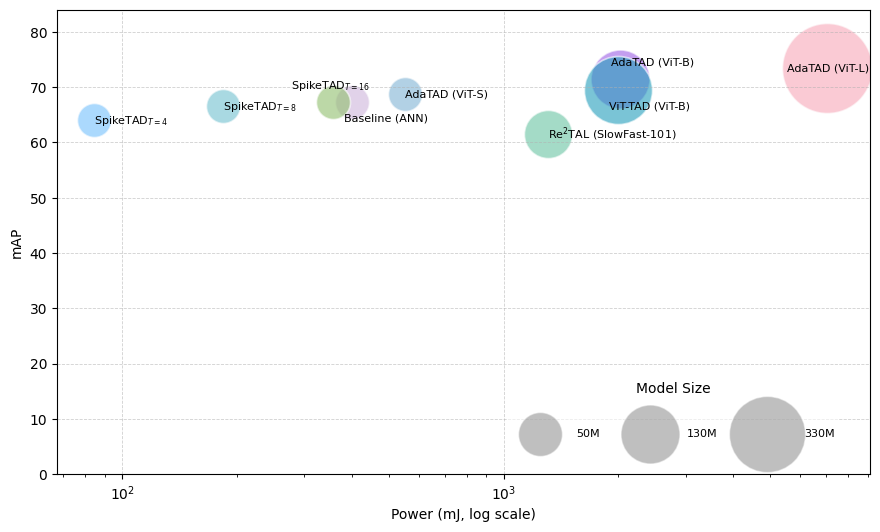}
	\caption{\textbf{Comparison between SpikeTAD and other methods.} We compared the energy consumption ratio and detection performance of SpikeTAD with other end-to-end TAD methods for THUMOS14~\cite{thumos14}. Due to the excessive power consumption of these ANN methods compared to SpikeTAD, we use log on the power consumption axis to visualize the power consumption gap between different methods.}
	\label{fig:spiketad_}
\end{figure}

To solve the problems above, we not only need to ensure the video understanding ability of SNN, but also need to use as few time-steps as possible to accumulate spikes.
Existing end-to-end TAD frameworks~\cite{vittad,adatad,tallformer,re2tal,basictad} usually consist of a backbone and a detector. The backbone~\cite{i3d,tsn,videomae_v2} relies on pre-trained weights from action classification datasets~\cite{k400,internvid}, while the detector can be optimized from scratch. 
Specifically, we first build and train an ANN version of the TAD model, including the backbone network and the detector in an end-to-end manner. Next, we adopt an ANN-SNN transformation strategy to convert the ANN model into an SNN model by transferring the trained weights. For the non-linear activation units containing training weights in the backbone, we use a combination of Multi-Threshold Neuron and Expectation Compensation Module~\cite{mth,mth2} to maintain lossless information propagation as much as possible. For simple action detectors, we replace the continuous activation functions like ReLU with simple Integrate-and-Fire Neuron. To use fewer time steps and maintain performance as much as possible, we used a quantization clip-floor activation function~\cite{optimalannsnn} to eliminate transformation errors.

To this end, we build the first SNN-based framework for TAD, termed as \textbf{SpikeTAD}. 
We choose the ViT-S model~\cite{videomae_v2} for its high accuracy and low parameter count and computational cost.
We use a simple multi-scale feature pyramid~\cite{basictad,temporalmaxer} for its lighter architectures to construct multi-level features along with several convolutional blocks for detection in the detector. 
As shown in Fig.~\ref {fig:spiketad_}, our SpikeTAD achieves comparable results to many ANN methods with only a few time-steps, while consuming significantly less power. 
The main contributions of this work are: 

\begin{itemize}
	\item We are the first to explore the application of SNNs on TAD and propose the first SNN-based end-to-end TAD framework termed as \textbf{SpikeTAD}.
	\item SpikeTAD explores a variety of spiking neurons and chooses the most effective manner through detailed experimental exploration, achieving a balance between power consumption and detection accuracy.
	\item SpikeTAD achieves outstanding accuracy with low energy consumption on two widely-used TAD datasets, and achieves better detection performance than its ANN version.
\end{itemize}

\section{Related Work}

\subsection{Temporal Action Detection}
Temporal action detection (TAD) aims to localize the temporal interval of each action instance in an untrimmed video and recognize its action category. 
These methods can be divided into one-stage TAD and multi-stage TAD. One-stage TAD methods~\cite{actionformer,tadtr,basictad} aim to detect the boundaries and categories of action segments in a single shot, they offer efficiency but are constrained by default anchors, requiring meticulous hyperparameter tuning. Multi-stage TAD methods~\cite{bmn,afsd} often involve multiple stages to generate and refine action detection results. They usually generate action proposals using predefined temporal anchors, refining them via boundary regression and classification.
Due to limited GPU memory, most existing methods~\cite{actionformer,tadtr,bmn} use pre-extracted action recognition features as inputs to the TAD model, while others try to train an end-to-end model~\cite{basictad,afsd,tallformer}. 
In the past year, the TAD community has produced models with faster inference speed and fewer parameters~\cite{pbd} and better detection results~\cite{adatad}. However, no work has attempted to reduce the power consumption of TAD. Unlike the aforementioned TAD works that involve complex ANN structure optimization, our SpikeTAD attempts to introduce a low-power SNN architecture for TAD models, taking an important step towards deploying it on neuromorphic chips of mobile devices. 

\subsection{Spiking Neural Networks}
Spiking Neural Networks (SNNs)~\cite{snn,snnresnet} are a class of neural networks that more closely emulate the behavior of biological neurons compared to ANNs. The primary distinction lies in their use of discrete spikes for information transmission, as opposed to the continuous signals employed by ANNs. SNNs have two advantages.
The first is the addition of a time dimension, which is conducive to long-term learning without forgetting. The second one is that SNNs can transmit information in a binary manner, which is more energy-efficient. This makes SNNs ideal for edge computing and real-time applications involving spatiotemporal data. 
Our SpikeTAD is the first to apply SNN to TAD tasks, significantly reducing energy consumption while ensuring high detection performance.

\subsection{Learning Paradigms for Spiking Neural Networks}
There are currently two mainstream approaches to training SNNs.
The first paradigm is the conversion from ANN to SNN (ANN-SNN)~\cite{optimalannsnn,optimized,bridgingthegap,deepspikingneuralnetwork,reducingannsnn}, and the second one is the supervised learning~\cite{rectified,differentiable,reducinginformationloss,imloss,enofsnn}. The ANN-SNN conversion method involves initially training an ANN and subsequently converting it into a homogeneous SNN by transferring the trained weights and substituting non-linear activation neurons with spiking neurons.
However, being restricted to rate-coding, ANN-SNN conversion usually requires dozens or even hundreds of timesteps to obtain well-performed networks. Several works~\cite{optimalannsnn,optimized,mth,mth2} have tried to improve the rate-coding to minimize the conversion error and use fewer time-steps to solve these problems. On the other hand, supervised learning employs alternative functions during backpropagation to approximate the firing process, enabling the direct training of SNNs as if they were ANNs. For example, by regarding the SNN as a special RNN, a training method of back-propagation through time with
different kinds of surrogate gradient was proposed~\cite{surrogate}. Spike-Timing-Dependent Plasticity (STDP)~\cite{STDP} enabled pure SNNs to learn
temporal patterns effectively like the backpropagation algorithm in ANN, achieving impressive performance with just a few time-steps. 
Our SpikeTAD adopts the learning paradigms of ANN-SNN to maximize the use of backbone weights in ANN models. Despite this, we go deep into the errors caused by the ANN-SNN conversion and try our best to reduce the errors with fewer spikes.

\section{Method}
In this section, we first briefly review the preliminaries for SNNs, then introduce SpikeTAD and provide other details.

\subsection{Preliminary}

\subsubsection{Integrate-and-Fire Neuron}
The Integrate-and-Fire (IF)~\cite{optimalannsnn} neuron accumulates spikes within a fixed time-step $T$, which are then encoded into a membrane potential and sent to the next neuron.
If the IF neurons in the $l$-th layer receive the input $x^{l-1}(t)$ from the last layer at time-step $t$, the temporal potential of the IF neurons can be defined as:
\begin{align}
    m^{l}(t) &= v^{l}(t-1) + W^{l}x^{l-1}(t), \label{eq:mlt} \\
    s^{l}(t) &= H(m^{l}(t) - \theta^{l}), \label{eq:slt} \\
    x^{l}(t) &= \theta^{l}s^{l}(t), \label{eq:xlt} \\
    v^{l}(t) &= m^{l}(t) - x^{l}(t). \label{eq:vlt}
\end{align}
where $m^{l}(t)$ and $v^{l}(t)$ represent the membrane potential before and after the trigger of a spike at time-step $t$. $W^{l}$ denotes the synaptic weight in the $l$-th layer. When any element $m^{l}_{i}(t)$ from $m^{l}(t)$ exceeds the firing threshold $\theta^{l}$, the neuron will elicit a spike $s^{l}_{i}(t)$ and update the membrane potential $v^{l}_{i}(t)$, where $i$ is the index of the threshold. 
$\theta^{l}$ denotes the learnable firing threshold, which is often initialized or bounded by the maximum activation value of the layer to ensure optimal information transmission.
We use $H$ to represent the Heaviside step function, which equals 1 when the spike triggers, otherwise 0. At the same time, the spike $s^{l}(t)$ refers to the output spikes of all neurons in layer $l$ at time $t$. $x^{l}(t)$ is the postsynaptic potential of layer $l$, which equals $\theta^{l}$ if the neuron fires and 0 otherwise, to achieve spike intensity equivalence. We use the ``reset-by-subtraction" mechanism in Eq\eqref{eq:vlt} to avoid information loss.

\begin{figure}[ht]
	\centering
	\includegraphics[width=0.7\columnwidth]{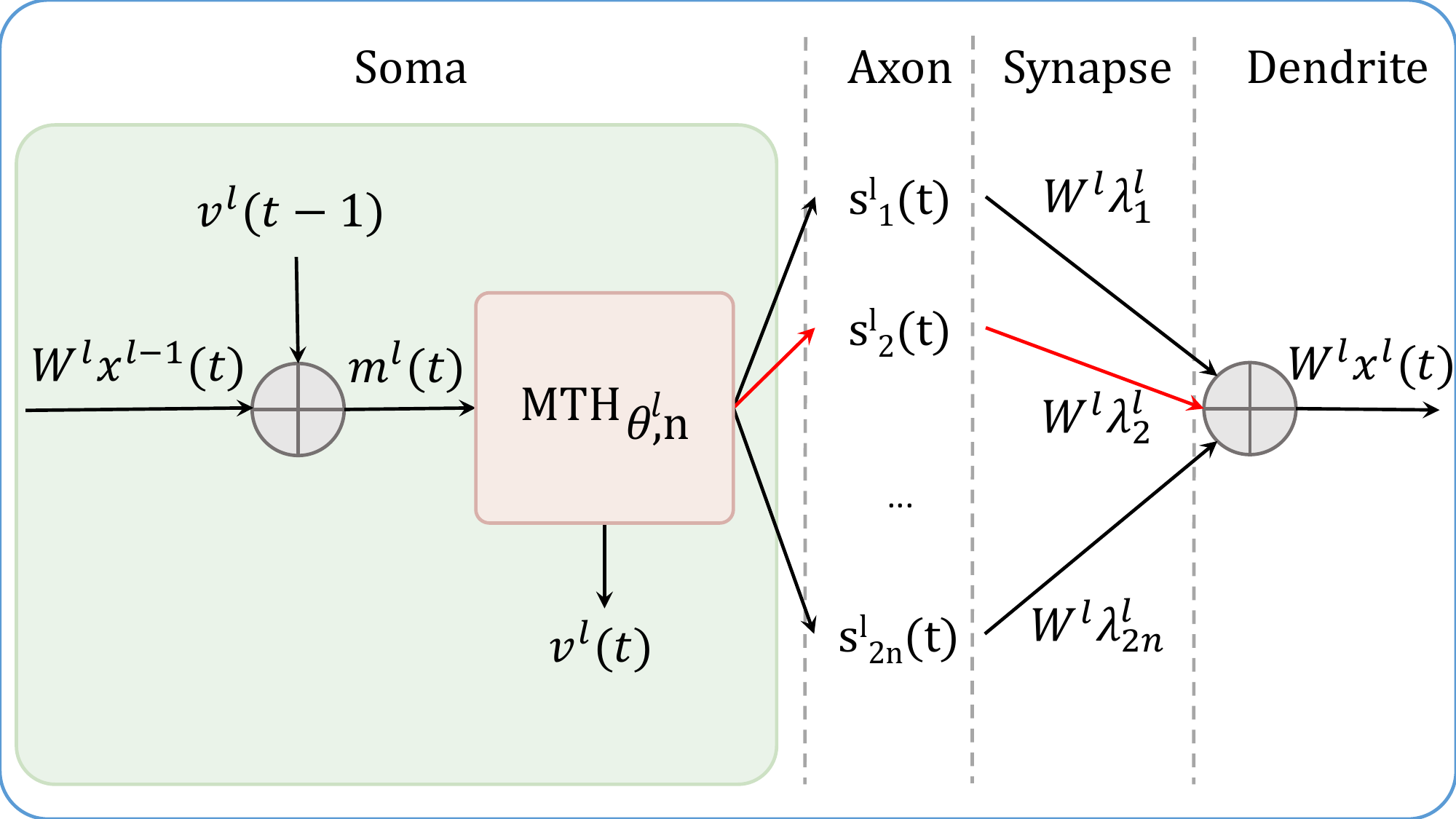}
	\caption{\textbf{Diagram of the multi-threshold neuron.} The multi-threshold neuron receives
input from last module and emits up to one spike at each time-step t.}
	\label{fig:mtn}
\end{figure}

\vspace{-4mm}
\subsubsection{Multi-Threshold Neuron}
IF neurons are mostly used to replace ReLU activation from linear or convolution layers in CNNs, 
but they cannot replace the activation operations like GELU from non-linear operations in Transformers. These operations require interaction between neurons within the same layer and exhibit non-linear characteristics. 
Integrate-and-Fire neurons can only send very simple spike signals and cannot express complex features. Forcing them to express complex features will result in a great loss of information, 
making it challenging to achieve accurate conversion through the linear piecewise quantization of individual neurons. 
To tackle this challenge, the multi-threshold neuron (MTN) is proposed.
MTN has a total of $2n$ thresholds with the base threshold $\theta$, with $n$ positive thresholds and $n$ negative thresholds. 
Both of them are learnable and derived from weights of the ANN function in the $l-$th layer. These thresholds divide the input signal into multiple intervals, thereby simulating more complex response patterns of neurons. 
Specifically, the MTN learning method describes the strength and polarity (positive or negative) of spike signals, rather than simply addressing the binary question of whether or not a spike is activated. 
However, MTN transitions from single-bit spike communication to multi-bit event representation, which inevitably increases the communication bandwidth and energy overhead compared to vanilla IF neurons.
Therefore, we only use MTN when we need to simulate spike activation in complex nonlinear layers within the backbone network.
We can refer to $\lambda^{l}_{i}$ as the $i$-th threshold value of MTN corresponding to index $i$.
\begin{align}
\lambda^{l}_{1} &= \theta^{l},\lambda^{l}_{2} = \frac{\theta^{l}}{2},\ \dots, \lambda^{l}_{n} = \frac{\theta^{l}}{2^{n-1}}, \label{eq:theta1} \\
\lambda^{l}_{n+1} &= -\theta^{l}, \lambda^{l}_{n+2} = -\frac{\theta^{l}}{2},\ \dots,\ \lambda^{l}_{2n} = -\frac{\theta^{l}}{2^{n-1}}. \label{eq:theta2}
\end{align}
Powers of 2 can give binary neurons multi-bit output capabilities and provide more fine-grained threshold representation.
As shown in Fig~\ref{fig:mtn}, the dynamics of MTN can be described as the following equations:
\begin{align}
m^{l}(t) &= v^{l}(t-1) + W^{l}x^{l-1}(t), \label{eq:mlt_m} \\
s^{l}_{i}(t) &= \text{MTH}_{\theta,n}(m^{l}(t),i), \label{eq:slit_m} \\
x^{l}(t) &= \sum_{i}s^{l}_{i}\lambda^{l}_{i}, \label{eq:xlt_m} \\
v^{l}(t) &= m^{l}(t) - x^{l}(t) \label{eq:vtl_m}, \\
\text{MTH}_{\theta,n}(m^{l}(t),i) &= \left\{
\begin{aligned}
0, & \quad \text{if } \lambda_{2n} < x <\lambda_{n}. \\
1, & \quad \text{elif} \quad i = \arg\min_{p} |x - \lambda_{p}|. \\
0, & \quad \text{else}.
\end{aligned}
\right. \label{eq:mth}
\end{align}

The only difference between MTN and IF is the generation of $s^{l}_{i}(t)$.
When $n = 1$, this model reduces to an IF neuron with an additional negative threshold.
 However, because MTN needs to transmit specific numerical values, the data packets transmitted between neurons become larger, significantly increasing the communication bandwidth burden and data exchange power consumption. Therefore, we only use it when necessary.

\subsubsection{ANN-SNN conversion}
For ANNs, the computations of analog neurons can be simplified as the
combination of a linear and a non-linear transformation:
\begin{align}
    a^{l} &= h(W^{l}a^{l-1}) = F(a^{l-1}), \label{eq:st_ann} 
\end{align}
Here $a^{l}$ denotes the output of all neurons in the $l$-th layer, $W^{l}$ denotes the weight of function $F$ in the $l$-th layer and $h(.)$ is a non-linear activation function. The key idea of ANN-SNN conversion is to map the activation value of a neuron in ANN to the firing rate of a spiking neuron in SNN. Specifically, we can get the potential update equation by combining equations from Eq~\eqref{eq:mlt} to Eq~\eqref{eq:vlt}. 
\begin{align}
    v^{l}(t) - v^{l}(t-1) &= W^{l}x^{l-1}(t) - s^{l}(t)\theta^{l}, \label{eq:vt_vt-1} 
\end{align}
By summing Eq~\eqref{eq:vt_vt-1} from 1 to T and dividing T on both sides, we obtain:
\begin{align}
    \frac{v^{l}(T) - v^{l}(0)} {T}&= \frac{W^{l}\sum^{T}_{i=1} x^{l-1}(i)}{T} - \frac{\sum^{T}_{i=1}s^{l}(i)\theta^{l}}{T}, \label{eq:vt_vt-1_T} 
\end{align}
We use $r^{l}(T)= \frac{\sum^{T}_{i=1} s^{l}(i)\theta^{l}}{T}$ to denote the encoded activation
that aims to map activation value $a^{l}$ from the corresponding ANN layer. When we substitute Eq~\eqref{eq:xlt} and Eq~\eqref{eq:vt_vt-1_T}, then we get:
\begin{align}
    r^{l}(T) &= W^{l}r^{l-1}(T) - \frac{v^{l}(T)-v^{l}(0)}{T}, \label{eq:rlt} 
\end{align}

Assuming the input rate $r^{l-1}(T) \approx a^{l-1}$, as the total time-steps $T$ increase or the residual term $\frac{v^{l}(T)-v^{l}(0)}{T}$ becomes negligible, the encoded rate $r^{l}(T)$ in the SNN asymptotically approximates the activation $a^{l}$ in the corresponding ANN. This mathematical equivalence ensures that the SNN can replicate the functional mapping of the ANN while operating with discrete spikes.

\subsubsection{Expectation Compensation Module}
In ANN-SNN conversion, using a limited number of time-steps ($T$) for efficiency often leads to severe information loss, particularly due to the non-linear nature of the modules. To mitigate this, the Expectation Compensation Module (ECM)~\cite{mth,mth2} leverages prior cumulative information to compute precise expectations and preserve non-linearity. For a non-linear layer $l$ with activation function $F$, let $S^{l-1}(T) = \sum_{t=1}^{T} x^{l-1}(t)$ be the cumulative input spikes up to time $T$. The exact spike output $x^l(T)$ at step $T$ can be deduced by subtracting the total expected output at $T-1$ from that at $T$:
\begin{align}
    x^{l}(T)
    &=TF(\frac{S^{l-1}(T)}{T})-(T-1)F(\frac{S^{l-1}(T-1)}{T-1}),\label{eq:xlT}
\end{align}
Eq~\eqref{eq:xlT} demonstrates that lossless conversion can be achieved simply by maintaining an accumulator for $S^{l-1}(T)$ from the previous layer. When extending this mechanism to matrix product operations—which are widely adopted in self-attention mechanisms—we apply the same expectation logic. Consider two spike input matrices, $A(t)$ and $B(t)$, with their respective cumulative sums denoted as $S_A(T) = \sum_{t=1}^{T} A(t)$ and $S_B(T) = \sum_{t=1}^{T} B(t)$. The cumulative matrix product $S_M(T) = S_A(T)S_B(T)$ can be updated recursively:
\begin{align}
    S_{M}(T)
    &= S_{M}(T-1)+A(T)B(T)+A(T)S_{B}(T-1)+S_{A}(T-1)B(T)  \label{eq:SM_update}
\end{align}
Consequently, the expected output matrix $x(T)$ at time $T$ is derived from the difference between consecutive expectation states:
\begin{align}
    x(T)
    &= \frac{1}{T}S_{M}(T)-\frac{1}{T-1}S_{M}(T-1)
\end{align}
This formulation indicates that a neuron requires only two variables to record the historical states $S_A(T-1)$ and $S_B(T-1)$. The main power consumption occurs during the computation of the update terms in Eq~\eqref{eq:SM_update}. However, because the input matrix guarantees limited effective thresholds per time-step, the total number of operations is highly constrained by the inherent spatial-temporal sparsity of the spikes. This allows the matrix product to be implemented through sparse accumulations, significantly reducing the energy footprint compared to dense ANN matrix multiplications.

\subsubsection{Quantization Clip-Floor Activation Function}
To mitigate performance degradation in ANN-SNN conversion at short time-steps, we replace the standard ReLU with a quantized clip-floor activation function~\cite{optimalannsnn,reducingannsnn,qffs}. To account for cases where the quantization steps $L$ differ from the SNN time-steps $T$, we introduce a shift parameter $\phi$. Let $z^l = W^l a^{l-1}$ denote the pre-activation potential. The shifted ANN activation function is defined as:
\begin{align}
    a^{l} &= \theta^{l}\text{clip}(\frac{1}{L}\lfloor \frac{z^{l}L}{\theta^{l}}+\phi \rfloor,0,1), \label{eq:al} 
\end{align}
In the corresponding SNN, accumulating $z^l$ over $T$ steps with an initial membrane potential $v^l(0)$ yields the expected firing rate $r^l(T)$. By tracking the membrane potential dynamics and bounding the residual potential $0 \le v^l(T) < \theta^l$, the firing rate can be directly expressed as:
\begin{align}
    r^{l}(T) &= \theta^{l}clip(\frac{1}{T}\lfloor \frac{z^{l}T+v^{l}(0)}{\theta^{l}} \rfloor,0,1) \label{eq:rlT_2} 
\end{align}
Our goal is to minimize the expected conversion error $\mathbb{E}_{z}[Err] = \mathbb{E}_{z}(r^{l}-a^{l})$. We can decouple this error into two components relative to the unquantized pre-activation $z^l$:
\begin{align}
    \mathbb{E}_{z}[Err] &= \mathbb{E}_{z}(r^{l}-z^{l}) + \mathbb{E}_{z}(z^{l}-a^{l}) \label{eq:err} 
\end{align}
To eliminate this error, we assume $z^l$ is locally uniform within small intervals. For a uniformly distributed random variable $x \in [0,\theta]$ with symmetrical boundary probability densities ($p_0 = p_T$), the expected quantization error with a half-step shift evaluates to zero:
\begin{equation}
\begin{split}
    \mathbb{E}_{x}\left(x-\frac{\theta}{T}\left\lfloor \frac{Tx}{\theta}+\frac{1}{2} \right\rfloor\right)
    &= (p_{0}-p_{T})\frac{\theta^2}{8T^2} = 0
\end{split}
\end{equation}
Applying this property to Eq~\eqref{eq:err}, we set the shift parameter $\phi = 1/2$ and the initial membrane potential $v^l(0) = \theta^l / 2$. This configuration independently zeroes out both terms ($\mathbb{E}_{z}(r^l - z^l) = 0$ and $\mathbb{E}_{z}(z^l - a^l) = 0$). Consequently, the mean conversion error strictly reaches zero ($\mathbb{E}_{z}[Err] = 0$), guaranteeing accurate ANN-SNN conversion even when the quantization steps $L$ are decoupled from the simulation time-steps $T$.

\subsubsection{Difference between Semantic Time and Computational Time}
\begin{figure}[ht]
	\centering
	\includegraphics[width=1.0\columnwidth]{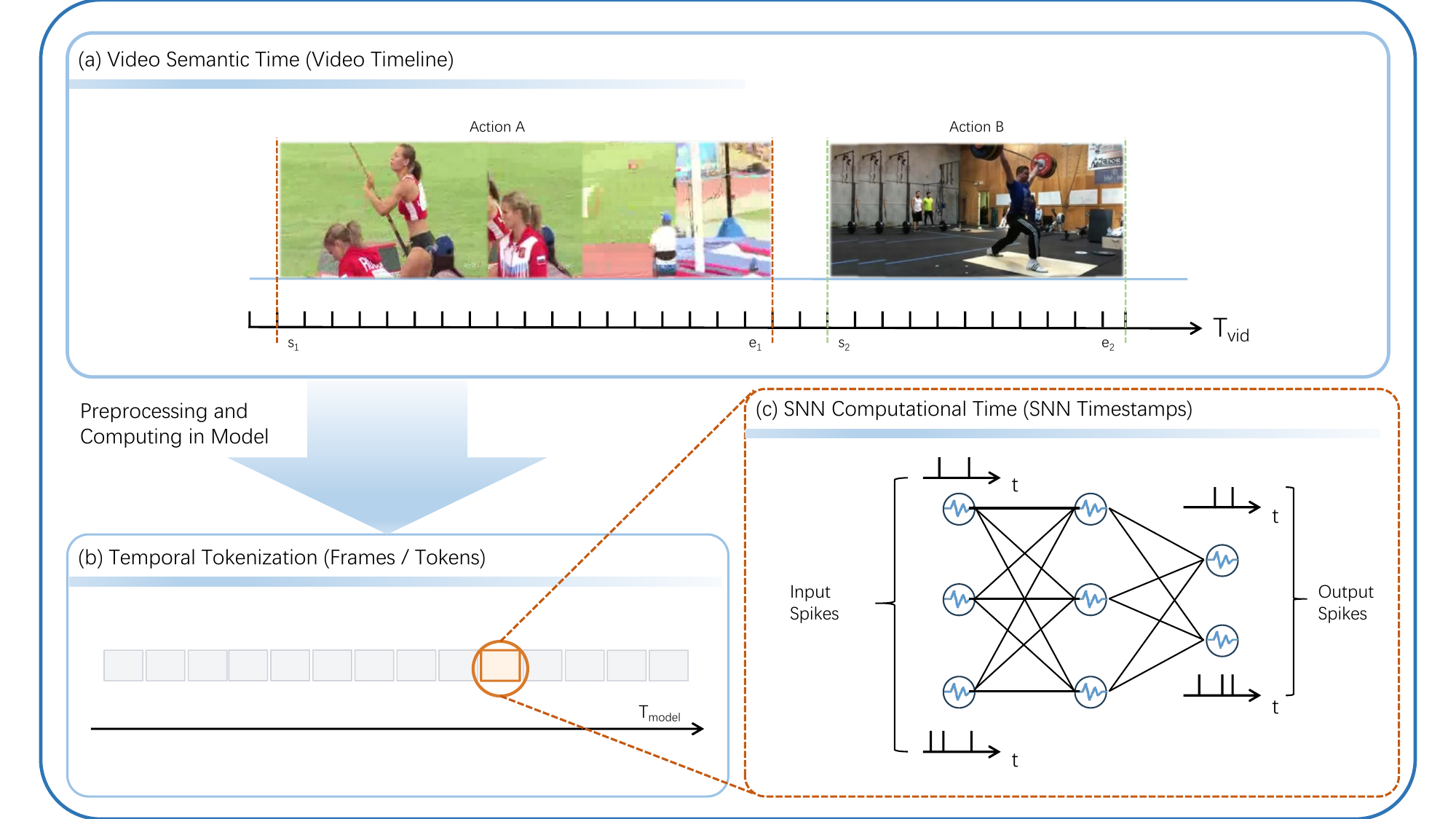}
	\caption{\textbf{Difference between semantic time and computational time.}}
	\label{fig:difference_t}
\end{figure}
As illustrated in Figure~\ref{fig:difference_t}, a fundamental distinction in our framework lies between the \textit{Semantic Time} (video frame index) and the \textit{Computational Time} (spike simulation steps). In video-based SNNs, the spike accumulation process operates on each individual frame, inherently leading to a multiplicative increase in computational overhead. To maintain high-precision ANN-SNN conversion within a strictly limited number of time-steps, we strategically deploy different neuronal mechanisms across the network. Specifically, for the three-dimensional spatio-temporal features ($T \times H \times W$) in the backbone, we employ the Expectation Compensation Module (ECM) and Multi-Threshold Neuron (MTN). The ECM preserves temporal fidelity by leveraging the statistical properties of mathematical expectation to compensate for residual information lost during quantization, thereby reducing the dependency on extensive time-steps. Simultaneously, the MTN enhances the information capacity per time-step and mitigates fine-grained quantization errors. For the detector stage, where features are compressed into the temporal-only dimension ($T$), we utilize a simplified Integrate-and-Fire (IF) neuron to ensure computational efficiency while maintaining effective spike activation.

\subsection{SpikeTAD Architecture}
\begin{figure*}[ht]
  \centering
  \includegraphics[width=0.95\textwidth]{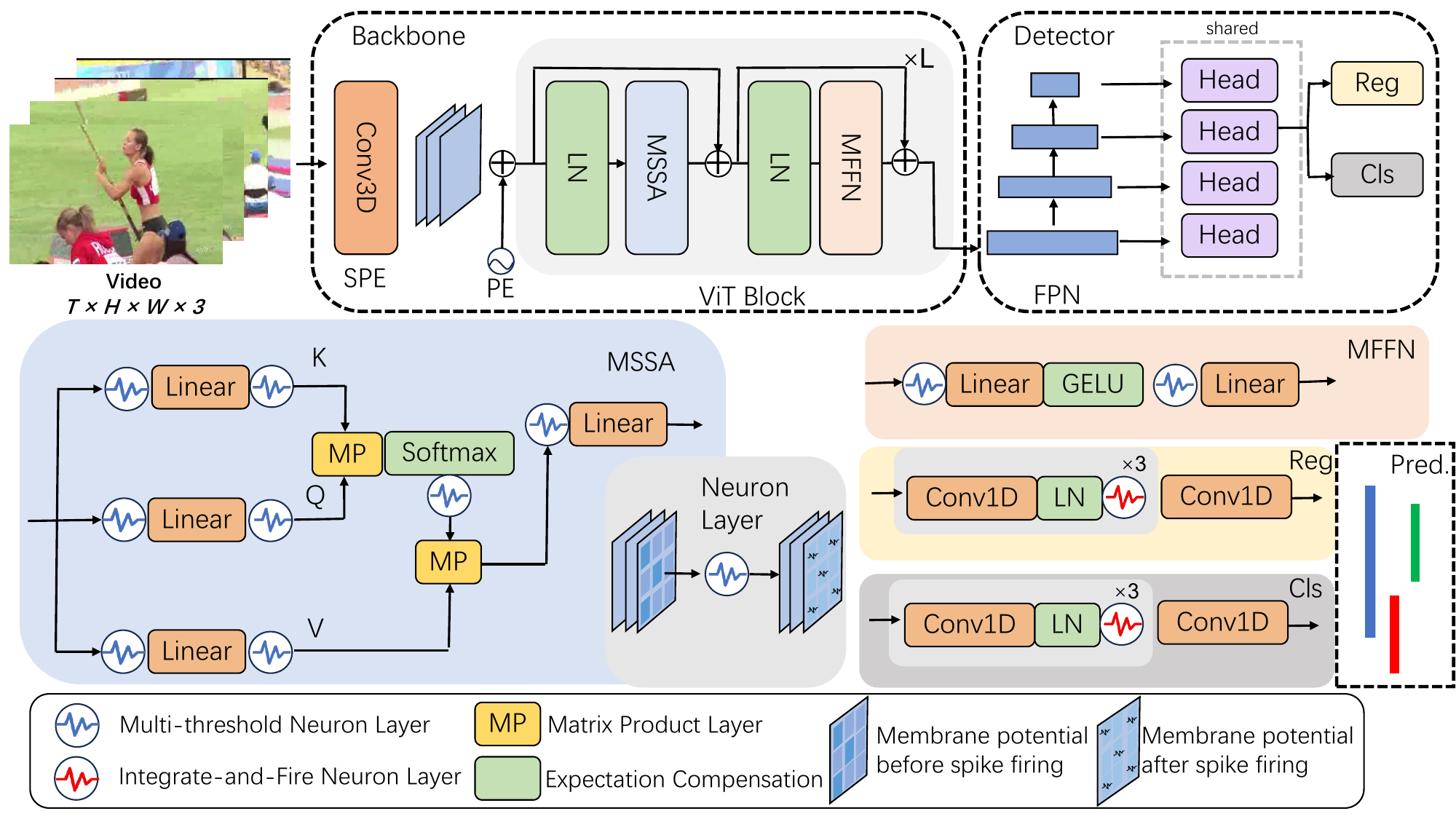}
  \caption{\textbf{The overview of SpikeTAD.} Our SpikeTAD consists of a backbone and detector. The backbone has $L$ ViT blocks. We transfer the input into spikes using multi-threshold neuron layer. We also transfer non-linear modules like GELU, LayerNorm and Softmax (marked in green) into corresponding expectation compensation modules to preserve prior information after each time-step. The detector adopts simple maxpooling layer to construct multi-scale features and uses integrate-and-fire neuron layer to replace the ReLU activation layer.}
  \label{fig:spiketad}
\end{figure*}
\subsubsection{Overview}

The overview of SpikeTAD is shown in Fig~\ref{fig:spiketad}. We input raw untrimmed videos with dimensions $T\times H \times W \times 3$ into our SpikeTAD. Then the input will be
split into patches and projected into tokens in spatiotemporal position embedding (SPE)~\cite{videomae_v2}. After adding the corresponding position embedding (PE), the spatiotemporal features will be input into the ViT block module with $L$ layers. We adopt expectation compensation module along with multi-threshold neuron to replace nonlinear operations in the ViT block. Specifically, each layer has Multi-Threshold Spiking Self-Attention (MSSA) and Multi-Threshold Feedforward network (MFFN) along with LayerNorm and shortcuts. The ANN's feature input $U$ needs to be transferred into postsynaptic potential features $X$ through spiking neurons by $X = \text{MTN}(U)$ before any Linear or matrix product operations. The whole procedure inside the $l-$th ViT block can be written as:
\begin{align}
    X^{l} &= \text{MSSA}(X^{l-1}) + X^{l-1}, l=1,...,L \label{eq:Xl}  \\
    X^{l} &= \text{MFFN}(X^{l}) + X^{l}, l=1,...,L \label{eq:Xl_2}
\end{align}

Specifically, the generation of $K,Q,V$ inside the MSSA can be represented as $X_{K,Q,V} = \text{MTN}(W_{K,Q,V}(\text{MTN}(X)))$. And 
 softmax and matrix product operations are as follows:
\begin{align}
 X^{l} &= \text{MP}(\text{MTN}(\text{Softmax}(\text{MP}(X^{l-1}_{K},X^{l-1}_{Q}))),X^{l-1}_{V}),  \label{eq:Xl_MSSA}
\end{align}
After channel dimension mapping $W_{c}$, we can obtain $X^{l}=W_{c}\text{MTN}(X^{l})$ as the output. We omit the relevant representation in MFFN which is similar to MSSA.
Then the feature $X$ in form of membrane potential will be input into the TAD's detector.
We construct multi-scale temporal features through the maxpooling layer and eliminate the spatial dimension of video feature, then feed them into the regression and classification branches for the final results. Specifically, we
add an Integrate-and-Fire Neuron layer to replace ReLU between the convolution operations:
    $X = \text{IF}(\text{LN}(\text{Conv1d}(X)))$.
The last convolution layer in the regression and classification branches will transform the postsynaptic potential features $X$ into the detection results of actions, in the form of triplet $(s_i,e_i,c)$ where $(s_i,e_i)$ represents the boundary of $i-$th action and $c$ is the label of category.

\subsubsection{ANN-SNN Conversion Module}
The ANN-SNN conversion uses rate coding,
which means the frequency of neural spikes per unit time, to transmit information. In this section, we will explain how neural spikes propagate information between nonlinear layers, including the mathematical derivation of expectation compensation modules and matrix products during inference within the backbone, and the quantization clip-floor-shift activation function used when training the ANN version of SpikeTAD inside the detector.
\paragraph{ANN-SNN Conversion inside Backbone}
In this section, We will present the transformation process of ANN-SNN in the backbone network of SpikeTAD using pseudocode.
\begin{algorithm}[htbp]
\caption{Operation in Multi-Threshold Neuron in MSSA and MFFN}
\label{alg:mtn}
\SetKwInOut{Input}{Input}\SetKwInOut{Output}{Output}
\Input{
    Input feature at time-step $t$ from layer $l$: $X^{l}(t)$\;
    Membrane potential at previous time-step $t-1$ from layer $l$: $V^{l}(t-1)$\;
    Number of positive/negative threshold $N$\; 
    Positive/Negative threshold value: $\theta_p,\theta_n$;
    Threshold scaling factor $g$\;
}
\Output{
    Accumulative spike tensor $S^{l}(t)$;
    Updated potential $V^{l}(t)$\;
}
\BlankLine
$V^{l}(t) \leftarrow V^{l}(t-1) + X^{l}(t)$;
$\lambda_p \leftarrow \theta_p \cdot g^{N-1}$;
$\lambda_n \leftarrow \theta_n \cdot g^{N-1}$;
$S^{l}(t) \leftarrow \mathbf{0}$;
$M_{mask} \leftarrow \mathbf{1}$
\BlankLine
\For{$i \leftarrow 1$ \KwTo $N$}{
    $H_p \leftarrow \text{Heaviside}(V^{l}(t) - \lambda_{p} + \theta_p / 2)$;
    $H_n \leftarrow \text{Heaviside}(-V^{l}(t) - \lambda_{n} + \theta_n / 2)$\;
    \BlankLine
    $T_p \leftarrow H_p \odot M_{mask}$;
    $T_n \leftarrow H_n \odot M_{mask}$\;
    \BlankLine
    $\Delta V \leftarrow (T_p \cdot \lambda_{p}) - (T_n \cdot \lambda_{n})$;
    $V^{l}(t) \leftarrow V^{l}(t) - \Delta V$;
    $S^{l}(t) \leftarrow S^{l}(t) + \Delta V$;
    \BlankLine
    $M_{mask} \leftarrow M_{mask} \odot (1-(T_{p} + T_{n}))$;
    $\lambda_{p} \leftarrow \lambda_{p} / g$;
    $\lambda_{n} \leftarrow \lambda_{n} / g$\;
}
\Return $S^{l}(t)$\;
\end{algorithm}
Algorithm~\ref{alg:mtn} details the operational dynamics of the Multi-Threshold Neuron (MTN) within the MSSA and MFFN modules. The process begins with the temporal integration of the membrane potential $V^{l}(t)$. To achieve high-fidelity quantization, we implement an iterative firing mechanism across $N$ threshold levels. We introduce a bias term $\theta/2$ within the Heaviside function to facilitate "round-to-nearest" quantization, consistent with the error-minimization proof in Eq~\eqref{eq:err}. The unexcited mask $M_{mask}$ ensures a "Winner-Takes-All" logic across scales, where only the highest applicable threshold is triggered for each neuron per time-step. Upon firing, a "reset-by-subtraction" operation updates the residual potential, preserving information for subsequent cycles. Finally, the thresholds are geometrically scaled by factor $g$ to refine the representation in the next iteration.

\begin{algorithm}[htbp]
\caption{Expectation Compensation Module for Matrix Product}
\label{alg:ecm}
\SetKwInOut{Input}{Input}\SetKwInOut{Output}{Output}
\Input{
    Input matrix tensor at $t$ time-step $A_t, B_t \in \mathbb{R}^{B \times L \times D}$\;
}
\Output{
    Compensated correlation matrix $O_t$\;
}
\BlankLine
\If{$t = 0$}{
    $S_A \leftarrow A_t$;
    $S_B \leftarrow B_t$;
    $S_M \leftarrow A_t \otimes B_t$;
    $t \leftarrow 1$;
    \Return $S_M$\;
}
    \BlankLine
     $\Delta_M \leftarrow (A_t \otimes B_t) + (A_t \otimes S_B) + (S_A \otimes B_t)$;
    $S_M \leftarrow S_M + \Delta_M$\;
    \BlankLine
    $S_A \leftarrow S_A + A_t$;
    $S_B \leftarrow S_B + B_t$;
    $t \leftarrow t + 1$;
    $O_t \leftarrow \frac{\Delta_M}{t-1} - \frac{S_M}{t(t-1)} $\;
    \Return $O_t$\;
\end{algorithm}

Algorithm~\ref{alg:ecm} presents the operational flow of the Expectation Compensation Module (ECM) for matrix products, typically utilized in self-attention mechanisms. For the input tensors $A_t$ and $B_t$ at time-step $t$, where $L$ represents the flattened spatiotemporal token dimension, the module first initializes the state at $t=0$. For subsequent steps, the ECM avoids the massive memory overhead of storing historical raw spikes by adopting a recursive update strategy. Specifically, we update the cumulative product $S_M$ using the incremental term $\Delta_M$, as derived in Eq~\eqref{eq:SM_update}. By maintaining only the cumulative sum tensors $S_A$ and $S_B$, the module efficiently computes the compensated correlation matrix $O_t$, which rectifies the quantization-induced expectation bias in short time-steps.

\begin{algorithm}[htbp]
\caption{Expectation Compensation Module for Non-linear Function}
\label{alg:exp_comp_neuron}
\SetKwInOut{Input}{Input}\SetKwInOut{Output}{Output}
\Input{
    Input tensor at $t$ time-step $x_t$,
    Non-linear function $f(\cdot)$\;
}
\Output{
    Compensated correlation matrix $y_t$\;
}
\BlankLine
\If{$t = 0$}{
    $V_{last} \leftarrow \mathbf{0}$;
    $S \leftarrow x_t$
}
\Else{
    $V_{last} \leftarrow f(S / t) \times t$;
    $S \leftarrow S + x_t$\;
}
$t \leftarrow t + 1$;
$V_{now} \leftarrow f(S / t) \times t$;
$y_t \leftarrow V_{now} - V_{last}$\;
\Return $y_t$\;
\end{algorithm}

Algorithm~\ref{alg:exp_comp_neuron} illustrates the ECM for non-linear functions (e.g., Softmax, GeLU, LayerNorm). Standard SNN quantization typically disrupts the precise probability distributions or normalization scales required by these operators. By maintaining a cumulative sum $S$ and computing the incremental contribution through the difference of expectations (i.e., $V_{now} - V_{last}$), the ECM ensures that the total accumulated output at any time-step $T$ remains functionally equivalent to the non-linear mapping of the input mean sequence. This differentially encoded output effectively captures transient semantic transitions between video frames, preserving high information fidelity even with minimal temporal latency.

\paragraph{ANN-SNN Conversion inside Detector} 
\begin{algorithm}[htbp]
\caption{Operation in Integrate-and-Fire Neuron in Head}
\label{alg:if_neuron}
\SetKwInOut{Input}{Input}\SetKwInOut{Output}{Output}
\Input{
    Input feature at layer $l$: $X^{l} \in \mathbb{R}^{B \times L \times C}$\;
    Accumulative time-step $T$;
    Learnable threshold $\theta$\;
}
\Output{
    Accumulative spike tensor $S_{acc}$\;
}
$X_{seq} \leftarrow \text{ExpandTemporalDim}(X^{l}, T)$;
$V^{l} \leftarrow 0.5 \cdot \theta$;
$S_{list} \leftarrow []$\;
\BlankLine
\For{$t \leftarrow 1$ \KwTo $T$}{
    \BlankLine
    $V^{l} \leftarrow V^{l} + X_{seq}[t, \dots]$;
    \BlankLine
    $s^{l} \leftarrow \text{Heaviside}(V^{l} - \theta)$;
    \BlankLine
    \BlankLine
    $X_{spike} \leftarrow s^{l} \cdot \theta$;
    $V^{l} \leftarrow V^{l} - X_{spike}$\;
    $S_{list}.\text{append}(X_{spike})$\;
}
\BlankLine
$S_{acc} \leftarrow \text{Stack}(S_{list}, \text{dim}=0)$\;
$S_{acc} \leftarrow \text{MeanTemporalDim}(S_{acc})$
\Return $S_{acc}$\;
\end{algorithm}

Algorithm~\ref{alg:if_neuron} illustrates the ANN-SNN conversion process within the detector head of SpikeTAD. For an input feature $X^l \in \mathbb{R}^{B \times L \times C}$, where $L$ denotes the temporal dimension of video features and $C$ represents the channel capacity, the module first replicates the features across $T$ simulation time-steps. To minimize the quantization-induced expectation bias, the membrane potential is initialized to $\theta/2$, consistent with our theoretical derivation in Eq~\eqref{eq:err}. During each time-step, the neuron performs temporal integration, spike firing via the Heaviside function, and a precise "reset-by-subtraction" operation. Finally, the generated spikes are accumulated and averaged to obtain the equivalent SNN activation, ensuring a seamless and high-fidelity transition from the ANN backbone to the spiking detector.

During inference, we simply substitute the function above into Integrate-and-Fire neurons due to the sparsity of spikes. Specifically, we set $m^{l}(0)=\frac{\theta^{l}}{2}$. From Eq~\eqref{eq:mlt} to Eq~\eqref{eq:vlt}, we can get each output $x^{l}(t)$ at each time-step $t$, and then we can get the output of this layer by $r^{l}(T)=\frac{\sum^{T}_{i=1}x^{l}(i)}{T}$ to denote the encoded activation that aims to map activation value from the corresponding ANN layer.

\subsubsection{Training loss}
SpikeTAD adopts simple loss functions as used in many other TAD methods. We use focal~\cite{focal} loss $L_{cls}$ for classification and DIoU~\cite{diou} loss $L_{reg}$ for distance regression. The total loss can be represented as $L_{total}=L_{cls}+L_{reg}$.

\subsubsection{Energy consumption.}
\begin{algorithm}[H]
\caption{SOP for Linear Layers}
\label{alg:sop_linear}
\SetKwInOut{Input}{Input}\SetKwInOut{Output}{Output}
\Input{Input tensor $x \in \mathbb{R}^{B \times M}$; Linear layer weights $W \in \mathbb{R}^{M \times N}$\;}
\Output{Active operands $SOP_{act}$; Total theoretical operands $SOP_{total}$\;}
\BlankLine
$SOP_{act} \leftarrow \text{Sum}(\text{Linear}(\mathbb{I}(x \neq 0), \mathbf{1}_{W}, \mathbf{0}))$;
$SOP_{total} \leftarrow \text{Sum}(\text{Linear}(\mathbf{1}_{x}, \mathbf{1}_{W}, \mathbf{0}))$\;
\Return $SOP_{act}, SOP_{total}$\;
\end{algorithm}

\begin{algorithm}[H]
\caption{SOP for Matrix Product in Expectation Compensation Module}
\label{alg:sop_matrix}
\SetKwInOut{Input}{Input}\SetKwInOut{Output}{Output}
\Input{Input  $A \in \mathbb{R}^{L \times D}$, $B \in \mathbb{R}^{D \times L}$,Cumulative historical state $S_A, S_B \in \mathbb{R}^{L \times D}$}
\Output{Active operands $SOP_{act}$, Total theoretical operands $SOP_{total}$}

\BlankLine

$M_A, M_B \leftarrow \mathbb{I}(A \neq 0), \mathbb{I}(B \neq 0)$
$M_{S_A}, M_{S_B} \leftarrow \mathbb{I}(S_A \neq 0), \mathbb{I}(S_B \neq 0)$

\BlankLine
$Ops_1 \leftarrow \text{Sum}(M_A \otimes M_B)$
$Ops_2 \leftarrow \text{Sum}(M_A \otimes M_{S_B})$
$Ops_3 \leftarrow \text{Sum}(M_{S_A} \otimes M_B)$

\BlankLine
$SOP_{act} \leftarrow Ops_1 + Ops_2 + Ops_3$; $SOP_{total} \leftarrow \text{Sum}(\mathbf{1}_{A} \otimes \mathbf{1}_{B})$\;

\Return $SOP_{act}, SOP_{total}$\;
\end{algorithm}

\begin{algorithm}[H]
\caption{SOP for Neurons}
\label{alg:sop_neurons}
\SetKwInOut{Input}{Input}\SetKwInOut{Output}{Output}
\Input{List of multi-level output spikes of neurons $Y = [y_1, y_2, \dots, y_N]$}
\Output{Active operands $SOP_{act}$, Total theoretical operands $SOP_{total}$}

\BlankLine
$SOP_{total} \leftarrow \text{numel}(y_1)$;
$SOP_{act} \leftarrow 0$\;

\BlankLine
\ForEach{$y_i \in Y$}{
    $SOP_{act} \leftarrow SOP_{act} + \text{CountNonZero}(y_i)$;
    $SOP_{total} \leftarrow SOP_{total} + numel(y_i)$\;
}

\Return $SOP_{act}, SOP_{total}$\;
\end{algorithm}
We use synaptic operation (SOP)~\cite{sop} for SNN to represent the required basic operation numbers. With the sparsity of firing and the short simulation period, the number of SOPs per
layer of the network can be easily estimated. A multiply-accumulate computation (MAC) takes place per synaptic operation for the ANN. Specialized SNN hardware would perform an accumulated computation (AC) per synaptic operation only upon receiving an incoming spike. Specifically, the energy consumption ratio between the SNNs and the related ANN is:
\begin{align}
    \frac{E_{SNN}}{E_{ANN}} = \frac{MACs_{SNN}*E_{MAC}+ACs_{SNN}*E_{AC}}{MACs_{ANN}*E_{MAC}}, \label{eq:Esnn_Eann} 
\end{align}
We use $\frac{E_{SNN}}{E_{ANN}}$ to represent the energy ratio in the subsequent experiments. 
All operations assume a 32-bit floating-point
implementation on 45nm technology, where $E_{MAC}$ = 4.6pJ and $E_{AC}$ = 0.9pJ~\cite{46pJ}. Since most computations in SNN occur in the fully connected, convolutional, and matrix multiplication layers, which are primarily implemented by additions (with $ACs_{SNN}$ \( >> \) $MACs_{SNN}$), we approximate $MACs_{SNN}$ to zero.
To be specific, the calculation of $ACs_{SNN}$ can be denoted as:
\begin{align}
    ACs_{SNN} =  T \times fr \times O_{AC},\label{eq:ACs_SNN} 
\end{align}
where T and $fr$ represent the simulation time-steps and the firing rate. The firing rate is the number of spikes released by each neuron divided by the total time-step T. 
The pseudocode computation process in SpikeTAD can be found in Algorithms \ref{alg:sop_linear}, \ref{alg:sop_matrix}, and \ref{alg:sop_neurons}. In these algorithms, we obtain the number of spurious activation units $SOP_{act}$ and the total number of operations $SOP_{total}$. Here, $fr$ is obtained by dividing $SOP_{act}/SOP_{total}$.
$O_{AC}$ is the total number of AC operations which can obtained by open-sourced calculations tool~\cite{syops}.

\section{Experiment}

\subsection{Datasets}
We evaluate SpikeTAD on two standard datasets for TAD: THUMOS-14~\cite{thumos14} and ActivityNet-1.3~\cite{anet}. \textbf{THUMOS14} is a classic dataset for the TAD task, which contains 413 videos of 30 fps, containing 200 validation videos, and 213 test videos with labeled temporal annotations from 20 categories. \textbf{ActivityNet-1.3} has 200 action classes. We use
the 10,024 videos from the training set for training and use
the 4,926 videos from the validation set for testing.

\subsection{Implementation Details}
We train and test SpikeTAD by feeding raw RGB videos, which means SpikeTAD is an end-to-end TAD pipeline. On THUMOS14, we randomly truncate a window of 768 frames with a temporal stride of 4. On ActivityNet-1.3, we resize the video to a fixed length of 768 frames. Following~\cite{actionformer,adatad}, we use AdamW~\cite{adamw} with warm-up for training.
The learning rate is 1e-4 for THUMOS14 and 1e-3 for ActivityNet-1.3, and a cosine learning rate decay is used. The batch size is 2 for THUMOS14 and 16 for ActivityNet-1.3. THUMOS14 takes 60 epochs for training while ActivityNet-1.3 takes 15 epochs. The random seed is 42 for reproducing our model results.

\subsubsection{Data processing}
During training, each video only randomly samples 768 consecutive frames, with the starting point of the sampling window being random. Then, when spatial changes occur on the video frames, the shorter side of the video is uniformly scaled to 182 pixels to allow space for subsequent cropping. Random cropping is then performed, randomly selecting a region from the $182 \times 182$ area to simulate random camera movement and scaling effects. The cropped area fluctuates randomly between 90\% and 100\% of the original image. The randomly cropped area is then forcibly stretched or compressed to a final $160 \times 160$ size. To ensure the model recognizes more than just "color" and "direction," a 50\% probability horizontal flip, image enhancement (including basic image processing such as blurring, noise reduction, and sharpening), and color jitter are introduced.
During testing, each video was sampled using a 768-frame sliding window. The Decord library was then used for video decoding, followed by spatial transformation of the captured video frames. The shorter side of the video was scaled to 160 pixels, and the scaled image frames were then cropped from the center to a $160 \times 160$ pixel size. These $768 \times 160 \times 160$ video frames were then input into SpikeTAD for inference.

\subsubsection{Hyperparameter table}
The hyperparameters used in the above algorithm and the statistics of random seeds used in training the code are shown in Table~\ref{tab:hyperparams}.

\begin{table}[htbp]
    \caption{Hyperparameter Configurations}
    \label{tab:hyperparams}
    \small 
    \begin{tabularx}{\linewidth}{@{} X c @{\hspace{1em}} X c @{}}
        \toprule
        \textbf{Hyperparameter} & \textbf{Value} & \textbf{Hyperparameter} & \textbf{Value} \\ 
        \midrule
        Random Seed  & 42 & Scaling Factor ($g$)     & 2 \\
        Time Steps ($T$)          & 16 & Quantization Step ($L$)   & 8 \\
        Base Threshold Positive ($\theta_p$) in Softmax & 0.0125 & Base Threshold Negative ($\theta_n$) in Softmax & 0.0125 \\
        Base Threshold Positive ($\theta_p$) in Linear & 0.25 & Base Threshold Negative ($\theta_n$) in Linear & 0.08 \\
        Number of Threshold in MTN ($N$)          & 8 &    &  \\
        \bottomrule
    \end{tabularx}
\end{table}

\subsection{Comparison with SOTA Methods}
\begin{table*}[ht]
\caption{\textbf{Comparison with state-of-the-art methods on THUMOS14 and ActivityNet-1.3.} Baseline(ANN) represents the ANN structure of SpikeTAD equipped with ReLU activation function. $T$ represents the time-steps in the backbone.}
\label{table:sota}
\centering
\resizebox{1.0\textwidth}{!}{
\begin{tabular}{c|c|c|cccccc|cccc}
\hline
\multirow{2}{*}{Type} & \multirow{2}{*}{Method}  & \multirow{2}{*}{Backbone} & \multicolumn{6}{c}{THUMOS14} & \multicolumn{4}{c}{ActivityNet-1.3} \\ \cline{4-13} 
& & & 0.3 & 0.4 & 0.5 & 0.6 & 0.7 & Avg & 0.5 & 0.75 & 0.95 & Avg \\ \hline
\multirow{16}{*}{ANN} 
& BMN~\cite{bmn} & TSN~\cite{tsn} & 56.0 & 47.4 & 38.8 & 29.7 & 20.5 & 38.5 & 50.07 & 34.78 & 8.29 & 33.85 \\
& TadTR~\cite{tadtr} & I3D~\cite{i3d} & 62.4 & 57.4 & 49.2 & 37.8 & 26.3 & 46.6 & 49.10 & 32.60 & 8.50 & 32.30 \\
& ActionFormer~\cite{actionformer} & SlowFast-R50~\cite{slowfast} & 78.7 & 73.3 & 65.2 & 54.6 & 39.7 & 62.3 & 54.26 & 37.04 & 8.13 & 36.02 \\
& ActionFormer~\cite{actionformer} & I3D~\cite{i3d} & 82.1 & 77.8 & 71.0 & 59.4 & 43.9 & 66.8 & 53.50 & 36.20 & 8.20 & 35.60 \\
& Tridet~\cite{tridet} & I3D~\cite{i3d} & 83.6 & 80.1 & 72.9 & 62.4 & 47.4 & 69.3 & 54.70 & 38.00 & 8.40 & 36.80 \\
& TALLFormer~\cite{tallformer} & VideoSwin-B~\cite{videoswintransformer} & 76.0 & - & 63.2 & - & 34.5 & 59.2 & 54.10 & 36.20 & 7.90 & 35.60 \\
& BasicTAD~\cite{basictad} & SlowOnly-R50~\cite{slowfast} & 75.5 & 70.8 & 63.5 & 50.9 & 37.4 & 59.6 & 51.20 & 33.41 & 7.57 & 33.12 \\
& ViT-TAD~\cite{vittad} & ViT-S~\cite{videomae_v2} & 79.8 & 75.2 & 68.4 & 56.4 & 41.7 & 64.3 & 55.09 & 37.81 & 8.75 & 36.69 \\
& ViT-TAD~\cite{vittad} & ViT-B~\cite{videomae_v2} & 85.1 & 80.9 & 74.2 & 61.8 & 45.4 & 69.5 & 55.87 & 38.47 & 8.80 & 37.40 \\
& AdaTAD~\cite{adatad} & ViT-S~\cite{videomae_v2} & 84.5 & 80.2 & 71.6 & 60.9 & 46.9 & 68.8 & 56.15 & 38.99 & 9.07 & 37.85 \\
& AdaTAD~\cite{adatad} & ViT-B~\cite{videomae_v2} & 87.0 & 82.4 & 75.3 & 63.8 & 49.2 & 71.5 & 56.77 & 39.35 & 9.71 & 38.39 \\
& AdaTAD~\cite{adatad} & ViT-L~\cite{videomae_v2} & 87.7 & 84.1 & 76.7 & 66.4 & 52.4 & 73.5 & 57.69 & 40.56 & 10.13 & 39.22 \\
& Re\textsuperscript{2}TAL~\cite{re2tal} & Re\textsuperscript{2}VideoSwin-T~\cite{re2tal} & 77.0 & 71.5 & 62.4 & 49.7 & 36.3 & 59.4 & 54.75 & 37.81 & 9.03 & 36.80 \\
& Re\textsuperscript{2}TAL~\cite{re2tal} & Re\textsuperscript{2}SlowFast-101~\cite{re2tal} & 77.4 & 72.6 & 64.9 & 53.7 & 39.0 & 61.5 & 55.30 & 37.90 & 9.10 & 37.00 \\
& Baseline(ANN) & ViT-S~\cite{videomae_v2} & 82.7 & 77.9 & 69.8 & 60.2 & 46.0 & 67.3 & 56.09 & 38.71 & 8.86 & 37.64 \\  
& SpikeTAD(ANN) & ViT-S~\cite{videomae_v2} & 82.6 & 77.7 & 69.7 & 60.0 & 45.6 & 67.1 & 55.83 & 37.92 & 8.67 & 37.12 \\ 
\hline
\multirow{3}{*}{ANN2SNN} 
& SpikeTAD$_{T=4}$ & ViT-S~\cite{videomae_v2} & 79.0 & 74.6 & 67.0 & 57.2 & 42.3 & 64.0 & 55.26 & 37.21 & 8.80 & 36.53 \\
& SpikeTAD$_{T=8}$ & ViT-S~\cite{videomae_v2} & 81.4 & 76.6 & 70.0 & 60.0 & 44.8 & 66.5 & 55.87 & 37.99 & 8.69 & 37.05 \\    
& SpikeTAD$_{T=16}$ & ViT-S~\cite{videomae_v2} & 82.1 & 77.9 & 70.5 & 59.9 & 45.6 & 67.2 & 56.04 & 38.97 & 8.75 & 37.42 \\ \hline
\end{tabular}
}
\end{table*}

Table~\ref{table:sota} compares our SpikeTAD with other state-of-the-art methods on THUMOS14 and ActivityNet-1.3. Since we aim to make the architecture simple and the computational complexity of inference small, our results cannot be compared with SOTA methods such as AdaTAD. Despite this, the ANN version of SpikeTAD still outperforms most TAD tasks.
When the time-step is 8, the detection performance of SpikeTAD decreases by 0.9 and 0.65 compared with the baseline (ANN) on THUMOS14 and ActivityNet-1.3, respectively.
SpikeTAD achieves tiny performance loss on TAD tasks, further demonstrating that the conversion of ANN-SNN is suitable for the TAD task.

\subsection{Ablation Study}

\subsubsection{Ablations on different SNN learning paradigms for TAD}
Since there are currently no SNN architectures specifically designed for the TAD task, this section examines the impact of various SNN paradigms on TAD performance. Most existing SNN architectures for object detection tasks~\cite{spikeyolo, spikingyolo} primarily adopt fully supervised training method to  directly train SNN models. Here, we refer to the approach in~\cite{spikeyolo} to build a pure SNN-based TAD model. Subsequently, we apply Integrate-and-Fire neurons directly to all activation functions, using clip-floor shift activation functions during training and accumulating spikes during testing to obtain detection results, denoted as ANN2SNN*. Finally, our proposed ANN2SNN paradigm incorporates an expectation compensation module with multi-threshold neurons in the backbone and employs clip-floor shift activation functions in the detector. The results are presented in Table~\ref{table:pure_snn}. We observed that both the SNN and ANN2SNN* approaches yielded comparable results, as they indiscriminately replaced activation units in the Backbone with their respective spiking neurons, leading to significant performance degradation. In contrast, ANN2SNN preserves the pre-trained weights of the Backbone, achieving performance close to the ANN model at T=8 time steps and even surpassed the ANN at T=16.

\begin{table}[ht]
\caption{\textbf{Comparison of different neural networks for TAD.} ANN is treated as baseline. SNN means Leaky Integrate-and-Fire spiking neurons that combine
integer-valued training with spike-driven inference. ANN2SNN* means clip-floor shift activation functions replace all activation functions like ReLU and GELU during training. ANN2SNN means our choice described in ANN-SNN Conversion Module.}
\label{table:pure_snn}
\centering
\resizebox{\columnwidth}{!}{
\begin{tabular}{c|c|c|ccccccc}
\hline
Model&Type & Time-step T & 0.3  & 0.4  & 0.5  & 0.6  & 0.7  & Avg.mAP \\ \hline
 \multirow{13}{*}{Backbone+Detector}&  ANN & -&82.6 & 77.7 & 69.7 & 60.0 & 45.6 & 67.1 \\ \cline{2-9} 
  & \multirow{4}{*}{SNN}   & 2           & 76.4 & 71.2 & 64.3 & 53.1 & 38.7 & 60.7    \\
&  & 4           & 77.7 & 72.1 & 64.5 & 54.2 & 39.5 & 61.6    \\
&     & 8           & 78.7 & 73.0 & 65.1 & 54.7 & 40.4 & 62.4    \\ 
&     & 16           & 78.8 & 72.9 & 65.2 & 54.8 & 40.5 & 62.5    \\ 
\cline{2-9}
 & \multirow{4}{*}{ANN2SNN*}   & 2           & 58.1 & 53.2 & 46.5 & 36.7 & 22.7 & 43.4    \\
& & 4           & 77.0 & 71.9 & 63.2 & 54.0 & 39.4 & 61.1    \\
&  & 8           & 78.8 & 72.6 & 64.5 & 53.6 & 39.4 & 61.8    \\
&     & 16           & 78.7 & 72.7 & 64.3 & 54.4 & 40.6 & 62.1    \\ \cline{2-9}
&  \multirow{4}{*}{ANN2SNN}  & 2& 64.5 & 58.8 & 51.7 &43.3 &32.3 &50.2 \\
& & 4& 79.0 & 74.6 & 67.0 &57.2 &42.3 &64.0 \\
& & 8 & 81.4 & 76.6 & 70.0 & 60.0 & 44.8 & 66.5 \\
& & 16 & 82.1 & 77.9 & 70.5 & 59.9 & 45.6 & 67.2 \\ \hline
\end{tabular}
}
\end{table}

\begin{table}[ht]
\caption{\textbf{Comparison between clip-floor shift with baseline clip-floor function.} The baseline takes more time-steps to simulate and the performance degrades a lot.}
\label{table:detector_choice}
\centering
\resizebox{\columnwidth}{!}{
\begin{tabular}{c|c|c|c|cccccc}
\hline
Model  &Type  & Func. & Time-step T & 0.3 & 0.4 & 0.5 & 0.6 & 0.7 & Avg.mAP \\ \hline
\multirow{7}{*}{Detector} & \multirow{7}{*}{ANN2SNN} & \multirow{6}{*}{Clip-Floor}         & 8         &  42.3   &  20.8   &   5.1  &   1.2  & 0.3    &  13.9       \\
    &     &                       & 16        &  70.1   &  63.7   &  51.6   &   32.4  &   8.9  &    45.4     \\
    &     &                       & 32        &   74.8  &  68.9   &   61.5  &  49.4   &  30.3   &    56.9     \\
    &     &                       & 64        &   75.8  &  70.8   &  63.4   &   53.0  &  36.6   &  60.0       \\
    &     &                       & 128       &   76.4  &  70.9   &  63.7   &  54.0   &  38.6   &   60.7      \\
    &     &                       & 256       &   76.5  &  71.1   & 63.6    &  53.8   &  39.5   &   60.9      \\ \cline{3-10} 
    &     & \multirow{1}{*}{Clip-Floor Shift}      & 4         &   82.1  &  77.7   & 70.7    &  60.1   &  46.0   &   67.3      \\ \hline  
\end{tabular}
}
\end{table}

\subsubsection{Effectiveness of Clip-floor Activation Function}
To further verify the effectiveness of clip-floor activation function,
we propose the baseline with the standard clip-floor activation function where we adopt $a^{l} = \theta^{l}\text{clip}(\lfloor \frac{W^{l}a^{l-1}}{\theta^{l}} \rfloor,0,1)$ at layer $l$ to substitute the ReLU activation function in the detector during training, which means the weights are normalized to between 0 and 1, and whether the signal is transmitted is determined by whether the spike is activated or not. During inference, we convert the models in both configurations into SNNs with spike activation. As shown in Table~\ref{table:detector_choice}, the baseline results still lag significantly behind the clip-floor shift activation function, even using more time-steps for simulation. Therefore, we adopt clip-floor shift activation function for the rest of our experiments. 

\begin{table}[ht]
\caption{\textbf{Detection results and energy consumption ratio of SpikeTAD's detector on THUMOS14.} ANN denotes the pure ANN version of the TAD baseline with ReLU. Energy ratio only calculates the modules in the detector.}
\label{table:energy_detector}
\centering
\resizebox{\columnwidth}{!}{
\begin{tabular}{c|c|c|c|cccccc}
\hline
Model & Type & Time-step T      & Energy Ratio & 0.3 & 0.4 & 0.5 & 0.6 & 0.7 & Avg. mAP \\ \hline
\multirow{6}{*}{Detector} &ANN          & - &     1.0         &  82.6 & 77.7 & 69.7 & 60.0 & 45.6 & 67.1\\ 
\cmidrule(r){2-10}
&\multirow{5}{*}{ANN2SNN}          & 1                &     0.03         &  80.6   &  76.5   &  67.7   & 55.4    &  41.4   &   64.3       \\
&             & 2                &     0.06         &  81.7   &  77.8   &   69.9  &  58.5   &  45.6   &    66.7      \\
&             & 4                &    0.14          &  82.0   &  77.8   &  70.4   &  60.0   &  45.8   &   67.2       \\
&             & 8               &    0.27          &  82.2   &  78.0   &  70.2   &  60.2   &  45.6   &  67.2        \\ 
&             & 16                &    0.55          &  82.1   &  77.7   &  70.7   &  60.1   &   46.0  &   67.3       \\ \hline
\end{tabular}
}
\end{table}

\begin{table}[ht]
\caption{\textbf{Detection results and energy consumption ratio of SpikeTAD's backbone on THUMOS14.} ANN denotes the pure ANN version of the TAD baseline with ReLU. Energy ratio calculates both the backbone and the detector.}
\label{table:energy_backbone}
\centering
\resizebox{\columnwidth}{!}{
\begin{tabular}{c|c|c|c|cccccc}
\hline
Model & Type & Time-step T & Energy Ratio & 0.3 & 0.4 & 0.5 & 0.6 & 0.7 & Avg. mAP \\ \hline
\multirow{6}{*}{Backbone+Detector} & ANN          & - &     1.0         &  82.6 & 77.7 & 69.7 & 60.0 & 45.6 & 67.1 \\
\cmidrule(r){2-10}
                          & \multirow{5}{*}{ANN2SNN} & 1 & 0.05  &  40.8   &  36.1   &   30.4  &  23.5   &  15.2   &  29.2        \\
                          &                      & 2 & 0.09  &  64.5   &  58.8   & 51.7    &  43.3   &  32.3   & 50.2         \\
                          &                      & 4 & 0.21  &  79.0   &  74.6   &  67.0   &  57.2   &  42.3   &   64.0       \\
                          &                      & 8 & 0.46  & 81.4    & 76.6    &  70.0   &  60.0   & 44.8    &   66.5       \\
                          &                      &16 & 0.89  &   82.1  &  77.9   &  70.5   & 59.9    &  45.6   &   67.2       \\ \hline
\end{tabular}
}
\end{table}

\begin{table*}[ht]
\caption{\textbf{Comparison between the SpikeTAD and previous end-to-end TAD models.} - means the model does not require or provide the corresponding parameters. 
}
\label{table:power_mJ}
\centering
\resizebox{1.0\textwidth}{!}{
\begin{tabular}{c|c|c|cccc}
\hline
Type & Method        & Backbone   & Time-Step & Param(M) & Power(mJ) \\  \hline
\multirow{9}{*}{ANN}   & TALLFormer~\cite{tallformer}        & VideoSwin-B~\cite{videoswintransformer}    &  -  & 114.77          &  8369.4        \\
             & ViT-TAD~\cite{vittad}          & ViT-B~\cite{videomae_v2}          &   -       &  131.27         &    1993.3      \\
             & AdaTAD~\cite{adatad}           & ViT-S~\cite{videomae_v2}          &  -        &  48.56         &   552.9      \\
             & AdaTAD~\cite{adatad}            & ViT-B~\cite{videomae_v2}          &   -       &  113.47         &  2024.6        \\
             & AdaTAD~\cite{adatad}           & ViT-L~\cite{videomae_v2}          &  -        &  331.44         &   7033.2       \\
             & Re$^2$TAL~\cite{re2tal}          & Re$^2$SlowFast-101~\cite{re2tal} &   -       &   91.52        & 1310.8         \\
             &Progressive Block Drop$_{block=3}$~\cite{pbd}          & ViT-S~\cite{videomae_v2} &   -       &   -        & 1002.8         \\
             & Baseline(ANN)    & ViT-S~\cite{videomae_v2}          &     -     &     48.58      &   400.4       \\ \hline
\multirow{3}{*}{ANN2SNN}          & \multirow{3}{*}{SpikeTAD}      &   \multirow{3}{*}{ViT-S~\cite{videomae_v2}}          &       4   &    48.58       &  84.1       \\ 
&       &           &       8   &    48.58       &      184.2   \\ 
&       &            &       16   &   48.58        &     356.4    \\ \hline
\end{tabular}
}
\end{table*}

\subsubsection{Energy consumption ratio of SpikeTAD.}

We can use open-sourced calculations tool~\cite{syops} to calculate the theoretical amount of AC and MAC operations. After that, we explore the power consumption and results when applying different time-steps in SpikeTAD's backbone and detector. As shown in Table~\ref{table:energy_detector} and Table~\ref{table:energy_backbone}, we find that SpikeTAD achieves its ANN version's performance when $T=8$, with the energy ratio of 46\% when calculating both backbone and detector. When $T=16$, the energy ratio is still lower than its ANN version.

\subsubsection{Energy consumption comparison with other end-to-end TAD methods}
The energy consumption of other end-to-end TAD models can be shown in Table~\ref{table:power_mJ}. We can find that
SpikeTAD's energy consumption is significantly lower than that of existing end-to-end TAD methods even the compressed TAD method~\cite{pbd} which costs over 1000 mJ. Both the low power consumption of neuromorphic chips and the low computational complexity of SpikeTAD contribute to such result.  Compare Baseline(ANN) with SpikeTAD(T=4), we can find that SpikeTAD's energy consumption is nearly one fifth of the Baseline. Since 45nm neuromorphic chips's $E_{AC}$ is close to one fifth of $E_{MAC}$, we can find that the total cost of neuron accumulated computations (AC) when $T=4$ still comparable to the number of MAC operations in ANNs. This indicates the sparse nature of SNN spike activations.

\begin{figure}[ht]
\centering
\includegraphics[width=1.0\columnwidth]{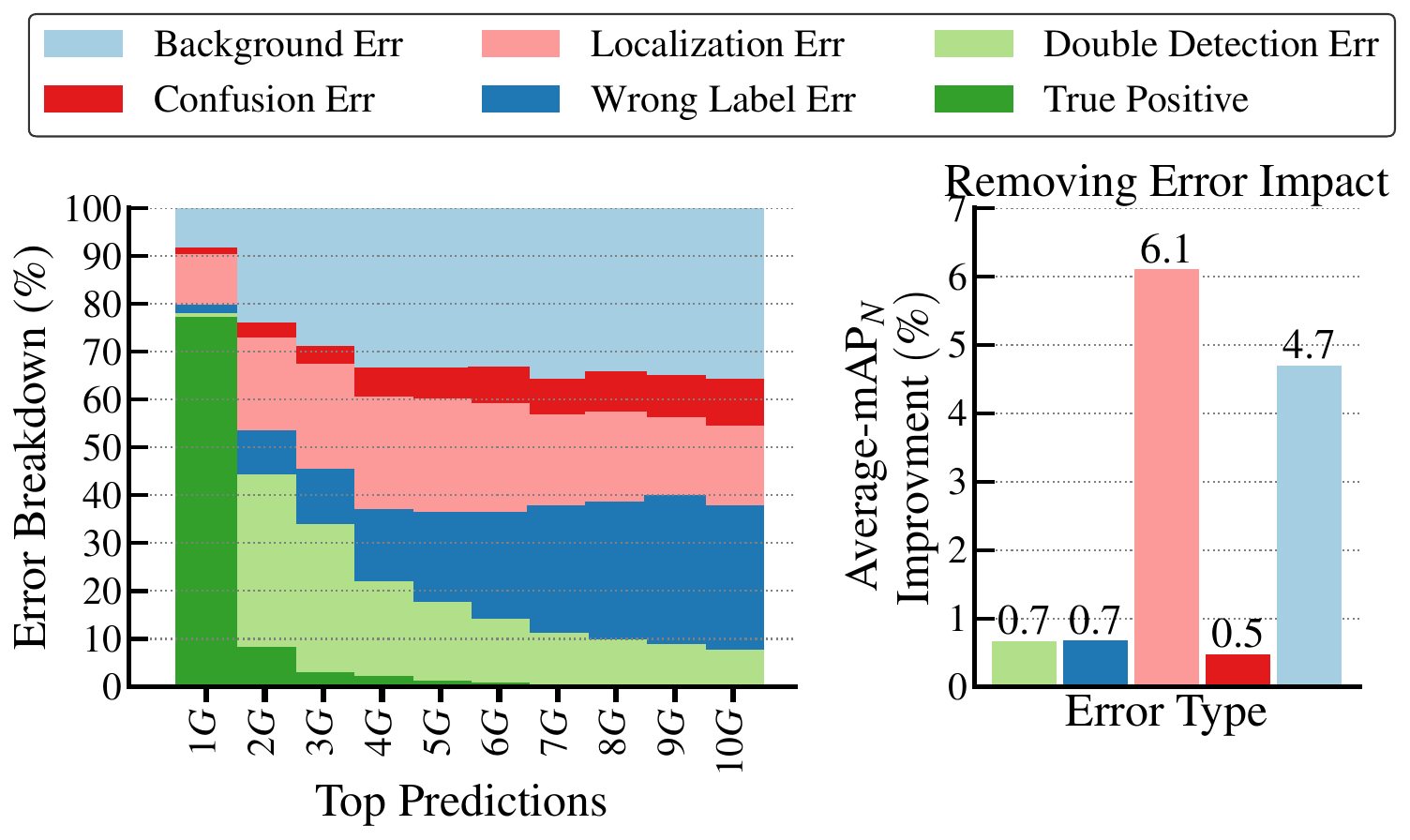}
	\caption{\textbf{Error analysis of SpikeTAD.} There are error rates of 5 types on top-10G predictions, where G denotes the number of ground truths.}
	\label{fig:error}
\end{figure}

\subsubsection{Statistical Significance}
\begin{table}[htbp]
    \centering
    \caption{\textbf{Statistical significance analysis of SpikeTAD over 5 independent runs.}}
    \label{table:statistical_significance}
    
    \setlength{\tabcolsep}{3pt}
    
    \resizebox{\linewidth}{!}{
        \begin{tabular}{@{} l ccccc c c @{}}
            \toprule
            \textbf{Metric} & \textbf{Run1} & \textbf{Run2} & \textbf{Run3} & \textbf{Run4} & \textbf{Run5} & \textbf{Mean$\pm$Std} & \textbf{95\% Conf. Inter.} \\ 
            \midrule
            mAP@0.3     & 82.1 & 82.3 & 82.4 & 81.8 & 81.9 & 82.10 $\pm$ 0.25 & [81.78, 82.42] \\
            mAP@0.4     & 77.9 & 77.8 & 78.0 & 77.7 & 77.6 & 77.80 $\pm$ 0.16 & [77.60, 78.00] \\
            mAP@0.5     & 70.5 & 70.2 & 70.5 & 70.2 & 70.4 & 70.36 $\pm$ 0.15 & [70.17, 70.55] \\
            mAP@0.6     & 59.9 & 60.1 & 60.2 & 59.7 & 60.0 & 59.98 $\pm$ 0.19 & [59.74, 60.22] \\
            mAP@0.7     & 45.6 & 45.5 & 45.6 & 45.4 & 45.8 & 45.58 $\pm$ 0.15 & [45.39, 45.77] \\ 
            \midrule
            Average mAP & 67.2 & 67.2 & 67.3 & 67.0 & 67.1 & 67.16 $\pm$ 0.11 & [67.02, 67.30] \\ 
            \bottomrule
        \end{tabular}
    }
\end{table}

For most deep learning models, the output hardly changes when all weights and hyperparameters, including the random seed, are fixed. Therefore, we further test the robustness of model training here. We train our ANN models with the following random seeds: 42, 123, 666, 1024, 2026, and apply the ANN2SNN technique from SpikeTAD, testing the results with an activation time step of 16. We calculate the mean $\mu$ and standard deviation $\sigma$ for the model results trained with different random seeds at different IoUs. Due to the small sample size, we use the t-distribution to calculate the confidence interval, with degrees of freedom $df=4$, a significance level $\alpha =0.05$, a two-sided critical value $t_{\alpha/2,df}=2.776$, and the confidence interval is $CI=\mu\pm t_{\alpha/2}\frac{\sigma}{\sqrt{n}}$. As shown in Table~\ref{table:statistical_significance}, SpikeTAD exhibits a remarkably low standard deviation, demonstrating that our proposed SNN-based temporal accumulation mechanism effectively suppresses initialization noise and ensures high training stability.

\subsubsection{Error Analysis} To analyze the limitations of our model, we provide false positive error chart~\cite{error} of our SpikeTAD on
THUMOS14 dataset shown in Fig~\ref{fig:error}. Background errors and localization errors are the most common. We obtain quite high true positive rate on Top-1G predictions.

\section{Limitations}
From a task-oriented perspective, the intrinsic complexity of Temporal Action Detection (TAD) necessitates a reliance on sophisticated pre-trained ANN backbones, which in turn elevates the difficulty of the ANN-to-SNN conversion process.
Specifically, two major challenges persist:
Computational Overhead: Due to the fundamental gap between semantic time and computational time, video-based SNNs encounter a significantly higher temporal redundancy than image-based models, as each video frame requires multiple internal spike accumulation steps.
Energy-Accuracy Trade-off: Unlike image processing where models can be trained from scratch (e.g., SpikingYOLO~\cite{spikingyolo}), the high-level semantic representation in TAD currently relies on converted ANN features. This forces the use of high-bit activation methods like the Multi-Threshold Neuron (MTN) to maintain competitive accuracy within constrained time-steps, leading to increased power consumption compared to binary spiking models.
Consequently, future research on SpikeTAD may follow two trajectories: developing end-to-end SNN architectures that support training-from-scratch to bypass the reliance on pre-trained ANN backbones, and optimizing the spike-firing mechanisms themselves to achieve high-fidelity conversion with lower power overhead.

\section{Conclusion}
In this paper, we propose the first SNN-based end-to-end TAD framework called SpikeTAD, which is simple enough and consumes much lower energy compared with other ANN-based TAD frameworks.
We adopt different ANN-SNN conversion strategies for the backbone and detector of TAD, so that the model can retain the pre-trained weights from the backbone as much as possible and minimize the conversion error. We hope that SpikeTAD’s successful attempt can trigger more thinking and attempts about low-power TAD architecture, enabling the deployment of TAD on hardware of neuromorphic computing. 

\textbf{Acknowledgments.}  
This work is supported by the Basic Research Program of Jiangsu (No. BK20250009), the Fundamental and Interdisciplinary Disciplines Breakthrough Plan of the Ministry of Education of China (No. JYB2025XDXM118), and the Collaborative Innovation Center of Novel Software Technology and Industrialization.

\appendix

\bibliographystyle{elsarticle-num} 
\bibliography{main}

\end{document}